\documentclass{ecai} 

\usepackage{latexsym}
\usepackage{amssymb}
\usepackage{amsmath}
\usepackage{amsthm}
\usepackage{booktabs}
\usepackage{enumitem}
\usepackage{graphicx}
\usepackage{color}

\usepackage{dsfont}
\usepackage{multirow}
\usepackage[dvipsnames]{xcolor}
\usepackage[noend]{algorithm2e}
\usepackage{caption}
\usepackage{subcaption}
\usepackage{float}
\usepackage{tikz}
\usepackage{hyperref}

\newtheorem{theorem}{Theorem}

\newtheorem{proposition}[theorem]{Proposition}

\newtheorem{definition}{Definition}

\newcommand{\BibTeX}{B\kern-.05em{\sc i\kern-.025em b}\kern-.08em\TeX}
\DeclareMathOperator*{\argmin}{arg\,min}
\DeclareMathOperator*{\argmax}{arg\,max}
\newcommand{\E}{\mathbb{E}}
\newcommand{\Var}{\mathbb{V}ar}

\begin{document}


\begin{frontmatter}


\paperid{123} 


\title{Fairness-Aware Grouping for Continuous Sensitive Variables: Application for Debiasing Face Analysis with respect to Skin Tone}



\author[A,B]{\fnms{Veronika}~\snm{Shilova}\thanks{Corresponding Author. Email: veronika.shilova@artefact.com}}
\author[A]{\fnms{Emmanuel}~\snm{Malherbe}}
\author[C]{\fnms{Giovanni}~\snm{Palma}}
\author[B]{\fnms{Laurent}~\snm{Risser}}
\author[B,D]{\fnms{Jean-Michel}~\snm{Loubes}}

\address[A]{Artefact Research Center, Paris, France}
\address[B]{Institut de Mathématiques de Toulouse (UMR 5219), CNRS, Université de Toulouse, F-31062 Toulouse, France}
\address[C]{L’Oréal Research and Innovation, Paris, France}
\address[D]{INRIA, ANITI, France}

%


\begin{abstract}
Within a legal framework, fairness in datasets and models is typically assessed by dividing observations into predefined groups and then computing fairness measures (e.g., Disparate Impact or Equality of Odds with respect to gender). However, when sensitive attributes such as skin color are continuous, dividing into default groups may overlook or obscure the discrimination experienced by certain minority subpopulations. To address this limitation, we propose a fairness-based grouping approach for continuous (possibly multidimensional) sensitive attributes. By grouping data according to observed levels of discrimination, our method identifies the partition that maximizes a novel criterion based on inter-group variance in discrimination, thereby isolating the most critical subgroups.

We validate the proposed approach using multiple synthetic datasets and demonstrate its robustness under changing population distributions—revealing how discrimination is manifested within the space of sensitive attributes. Furthermore, we examine a specialized setting of monotonic fairness for the case of skin color. Our empirical results on both CelebA and FFHQ, leveraging the skin tone as predicted by an industrial proprietary algorithm, show that the proposed segmentation uncovers more nuanced patterns of discrimination than previously reported, and that these findings remain stable across datasets for a given model.
Finally, we leverage our grouping model for debiasing purpose, aiming at predicting fair scores with group-by-group post-processing. The results demonstrate that our approach improves fairness while having minimal impact on accuracy, thus confirming our partition method and opening the door for industrial deployment.
\end{abstract}

\end{frontmatter}


\section{Introduction}
\label{sec:intro}

Group fairness measures inherently depend on the underlying distribution of machine learning data, such as specific input variables or the  predicted scores  for a given task. These measures are becoming increasingly impactful in light of recent AI regulations, such as the EU AI Act, which emphasize the need for reporting algorithmic performance across demographic groups.
Beyond reporting, addressing disparities generally involves mitigation strategies applied on a group-by-group basis, based
on qualitative sensitive variables. 
We refer, for instance, to \cite{
besse2018confidence, chouldechova2020snapshot, del2020review, dwork2012fairness, gordaliza2019obtaining, kamishima2011fairness, risser2022tackling, wang2022brief} and references therein.
The sensitive attribute can also be natively continuous, which is of primary interest in our paper. In this case, defining groups in the attribute space remains relevant, but raises the question on how to choose the partition and on which criteria.

A major field impacted by this problem is Computer Vision, for which widespread applications in high-stakes tasks have raised many ethical questions in recent years.
A very sensitive scenario is when models are applied to facial images, since the risk of bias with respect to (w.r.t.) skin tone might be related to ethnicity.
This risk may occur for models that address transversal problems, such as face verification and attribute prediction, as first highlighted by \cite{buolamwini2018gender} in which automated commercial facial analysis algorithms were shown to be biased against women with darker skin tones. It was later confirmed by \cite{thong2023beyond}, as well as for text-to-image generation models, including DALL-E \cite{cho2023dall}.
Such bias has been observed for many commercial applications, such as pedestrian detection \cite{li2023dark}, online ad delivery service \cite{sweeney2013discrimination}, or online cosmetic product recommendation \cite{malherbe2022skin}.
In this paper, we focus on the industrial scenario of a make-up recommendation algorithm based on selfie pictures.
The recommended products and make-up strategy should indeed only depend on user's face morphology and not on its skin tone nor ethnicity.

Although being central for these fairness topics, the effective way to measure skin tone remains an open question. Many studies \cite{buolamwini2018gender, schumann2024consensus} rely on manual annotators grouping individuals into predefined skin tone categories, such as the Fitzpatrick scale \cite{fitzpatrick1988validity} or Monk scale \cite{schumann2024consensus}. However, this approach presents three major drawbacks.
First, a simplistic division of tones risks oversimplifying the skin color and failing to capture nuanced biases embedded in datasets and models, which is obvious in the binary extreme of "darker" and "lighter" categories, but also occurs with the six groups of the commonly used Fitzpatrick scale.
Second, these categories are predefined, while fairness is always related to a given task. The discriminated groups thus differ from one task to another, and hence cannot be captured by a default grouping of individuals.
Third, manual annotations can introduce perception biases. Annotators might perceive skin tones differently depending on their cultural background \cite{garcia2016colored, gevers2012color}, which introduces annotator bias in the data.
To overcome the limitations described above, an alternative line of research \cite{robin2020beyond,thong2023beyond} focuses on the estimated skin tone as a continuous quantity expressed in the color space, thus capturing a broader spectrum of skin tones. However, in the context of fairness, the necessity to consider groups raises the question on how to segment such skin color space.

In this paper, we introduce a paradigm that tailors the partitioning process to a given classification task. 
Our contributions are:
\begin{itemize}
    \item We propose novel data-driven approaches to segment a continuous, possibly multidimensional sensitive variable into groups being homogeneous w.r.t. the discrimination present in a dataset or a model;
    \item We show through extensive experiments on skin tone that data-driven segmentation reveals more discrimination patterns than predefined groups and remains robust to population shifts, allowing us to better isolate how discrimination manifests within the sensitive attribute space. We also introduce and study the particular case of monotonic fairness, which may occur with skin tone;
    \item We use the groups identified by our data-driven method to mitigate the bias of a predictive model. Compared to other partitions, we obtain the best trade-off between model fairness and accuracy.
\end{itemize}

\section{Related work}
\textbf{Measures of fairness for groups and for continuous attributes}

Measuring and ensuring fairness in machine learning systems have become a central focus in recent years, leading to the development of various metrics assessing group-level disparities \cite{caton2024fairness}. Group fairness seeks to guarantee equitable treatment across predefined demographic groups, and several standard measures have been proposed to evaluate and mitigate biases in algorithmic decision-making.

The most commonly used measure of algorithmic bias is Statistical Parity \cite{corbett2017algorithmic}, frequently quantified in the fair learning literature through Disparate Impact (DI) \cite{feldman2015certifying}. Statistical Parity requires that individuals from different demographic groups have equal chances of receiving a positive outcome. However, its simplicity comes at a cost, as it does not account for the true distributions of target outcomes between groups. For instance, if one group is historically underrepresented in the positive class, targeting equal outcomes across groups reduces overall accuracy.  
%
On the other hand, the Equalized Odds \cite{berk2021fairness} metric focuses on ensuring that the rates of true positives and false positives are comparable across all groups. By considering the underlying differences between groups, this measure highlights error rate equality, making it particularly relevant in contexts where fairness in outcomes is critical, such as criminal justice and healthcare. Despite its advantages, achieving Equalized Odds in practice can be challenging without sacrificing predictive accuracy.

In the case of a continuous sensitive attribute, fairness is often quantified through a measure of statistical correlation. The most common measures are Hirschfeld-Gebelein-Rényi (HGR) maximum correlation coefficient \cite{mary2019fairness} and Generalized Disparate Impact \cite{giuliani2023generalized}, which is a modified formulation of HGR. These fairness metrics primarily evaluate the overall dependence between a continuous sensitive variable and a target variable, without specifying which groups within the population are more or less affected by discrimination. As a result, these metrics are not directly applicable to detect and locate biases in models. This limitation is particularly relevant in the context of emerging legal frameworks, such as the EU AI Act, which mandates that AI systems ensure they do not disadvantage specific groups based on attributes such as skin tone, age, or income, which, in fact, are inherently continuous.

\textbf{Skin color as a sensitive variable in Computer Vision}

While attributes representing skin tone is often provided in Computer Vision datasets, it is represented as a discrete attribute in the most commonly used ones. For example,
the skin color attribute has been annotated in the two groups \textit{Pale/Not Pale} in the CelebA dataset \cite{liu2015faceattributes},  in the 6 groups of the Fitzpatrick scale \cite{fitzpatrick1988validity} in the IJB-C \cite{maze2018iarpa} and Facet \cite{gustafson2023facet} datasets, and in the 10 groups of the Monk scale \cite{schumann2024consensus} in the MST-E dataset \cite{garcia2016colored}.
These commonly used groups for skin tone are of a limited number, when compared to the full range of skin colors. Besides, they are generally given by a human annotator (as in the datasets mentioned above), which has been observed to be biased with respect to its geographical region or its culture \cite{gevers2012color}, or to the inferred race/ethnicity of the individual in the picture \cite{garcia2016colored}.
%
To mitigate annotator bias and limited skin groups granularity, the recent line of work intended to consider skin tone as a measured color, thus a continuous quantity. The study \cite{thong2023beyond} proposed to extract skin color directly from pixels instead of relying on human perception and converted it in the CIELAB color space \cite{weatherall1992skin}. This color space, unlike RGB, correlates with the response of the human eye by covering its entire range of color perception and is a golden standard to measure skin color in clinical dermatology \cite{wilkes2015fitzpatrick} and aesthetic cosmetology \cite{ly2020research}. However,  estimation of skin color directly from pixels' color may be noisy as it depends on lightning conditions of a given picture, especially for in-the-wild pictures \cite{thong2023beyond}.
As a more reliable estimate, using a spectrophotometer to measure panelists' skin reflectance provides a ground truth value for training a supervised model \cite{ly2020research,robin2020beyond}. 

The previous approaches to quantifying fairness w.r.t. skin tone in datasets and models \cite{buolamwini2018gender,cho2023dall,li2023dark,thong2023beyond} predefine the sensitive attribute into binary or discrete groups. For instance, \cite{buolamwini2018gender,cho2023dall,li2023dark}, divide the population into two broad categories: \textit{Lighter} and \textit{Darker} skin tones. In contrast, \cite{thong2023beyond} proposes a more granular characterization with four subgroups that combine two dimensions: \textit{Lighter} and \textit{Darker} skin tones representing skin lightness, and \textit{Red} and \textit{Yellow} skin hues reflecting variations in perceived skin color gradation from reddish to yellowish tones. Further improvements can however be made, since these discretizations overlook both the underlying data distribution and the target variable. Another limit of these predefined groups is that samples inside a group are assumed to be homogeneous in terms of fairness, which is not guaranteed in reality. Besides, as already shown in \cite{kearns2018preventing}, overly simplistic group definitions may overlook fairness nuances and underestimate the level of discrimination present in the data or/and introduced by the model.

\section{Color space for skin tone}

In view of the litterature for skin tone estimation, in this work we consider the skin tone as a color in the three-dimensional CIELAB color space. We present below this color space and the color attributes commonly derived from it.
\begin{itemize}
\item {\textbf{CIELAB color space for measures.}}
CIE $L^*a^*b^*$ (CIELAB) color space \cite{seve2009science} was defined by the International Commission on Illumination in 1976  with the aim of approximating human perception of color differences. 
A color is expressed by a vector $(L^*,a^*,b^*) \in \mathbb{R}^3$ with the property that the Euclidean distance between two colors is proportional to their difference as perceived by humans, which is not verified in the RGB space. The $L^* \in [0, 100]$ component corresponds to the perceptual lightness and ranges from black to white. The $a^* \in [-128, 127]$ component refers to the green-red opponent, with negative values corresponding to green and positive values to red. The $b^* \in [-128, 127]$ component describes the blue-yellow opponent colors, with negative values corresponding to blue and positive values to yellow.

\item{\textbf{Individual typology angle.}}
The individual typology angle ($IT\!A$) is a one dimensional measure of skin tone defined by \cite{chardon1991skin} in the CIE $L^*a^*b^*$ color space as follows:
\begin{equation}
    IT\!A = arctan \bigg(\frac{L^* - 50}{b^*} \bigg) \times \frac{180^\circ}{\pi}.
\end{equation}

It is widely used for skin color measurements in dermatology \cite{wilkes2015fitzpatrick} and aesthetic cosmetology \cite{ly2020research}. In fairness field, $IT\!A$ is commonly applied to facial images \cite{karkkainen2021fairface, merler2019diversity} and skin analysis \cite{kinyanjui2020fairness}. The Fitzpatrick scale is derived from $IT\!A$ by predefined segmentation into 6 groups.

\item{\textbf{Hue angle.}}
Hue angle $h^*$ measures the perceived gradation of colors, and is defined as the angular coordinate in the $L^*a^*b^*$ space:
\begin{equation}
    h^* = arctan \bigg(\frac{b^*}{a^*} \bigg) \times \frac{180^\circ}{\pi},
\end{equation}

where $h^*$ ranges from $0^{\circ}$ to $360^{\circ}$. However, we are primarily interested in the values between $0^{\circ}$ to $90^{\circ}$, since they correspond to the red and yellow colors in which skin undertone is expressed \cite{chardon1991skin}. The hue angle was originally proposed for skin color measurements by \cite{weatherall1992skin}, and is now widely adapted for skin color estimation in cosmetology \cite{de2007development}.
\end{itemize}

In our experiments, to ensure that these quantities are not biased, we extract the skin tone values from face pictures leveraging a proprietary model implementing the methodology detailed in \cite{malherbe2022skin,robin2020beyond}. Since the underlying algorithm aims to approximate physical measurements from a device, it is not susceptible to human-induced annotation bias. The detailed methodology is beyond the scope of this paper.

\section{From continuous to discrete: enhancing fairness in sensitive variable segmentation}

In this paper, we place ourselves in the case of a dataset with a binary classification task as target outcome and a continuous sensitive attribute.
Formally, we consider a dataset consisting of i.i.d observations $(X, L, Y) \sim \mathcal{D}$, tuples of random variables $X \in \mathbb{R}^d$ ($d \in \mathbb{N})$, $L \in \mathcal{L} \subset \mathbb{R}^p$ ($p=1$ or $2$) and $Y \in \{0, 1\}$ drawn from a distribution $\mathcal{D}$. In our application, $X$ contains images of human faces in the RGB color space, $L$ is a continuous (multidimensional) sensitive attribute corresponding to skin color, such as $L^*$, $IT\!A$ or $(L^*, h^*)$ from CIE $L^*a^*b^*$ color space defined in previous section, and $Y$ is an observed outcome variable.
The parameters $\theta$ of a binary classifier $\hat{Y} = g_{\theta}(X) \in \{0,1\}$ are trained using the observations from the dataset $\mathcal{D}$.
In this setting, our objectives are two-fold:

\begin{enumerate}
    \item Showcase the relevance of using continuous (possibly, multidimensional) skin color score to measure bias in Computer Vision datasets and models;
    \item Propose a fairness-based segmentation of continuous (multidimensional) skin color. In other words, instead of measuring the fairness of a dataset or a model using predefined groups (e.g. Pale vs. Not Pale, Fitzpatrick groups), we propose to divide the observation according to groups in the skin color space $\mathcal{L}$. Such groups would need (i) to be connected and (ii) to have homogeneous levels of discrimination for the target problem $Y$ or model $\hat{Y}$.
\end{enumerate}

For the second objective, we formally write a partition $\mathcal{P} = \{\mathcal{P}_k\}_{k=1}^K$ as a set of connected subsets $\mathcal{P}_k \subset \mathcal{L}$ with the two following properties: there is no overlap between elements of the partition: $\forall k_1, k_2 \in \{1, \dots, K\}$, $\mathcal{P}_{k_1} \cap \mathcal{P}_{k_2} = \emptyset$; and $\mathcal{P}$ totally covers $\mathcal{L}$: $\cup_{k=1}^K \mathcal{P}_k = \mathcal{L}$. We consider that the number of groups $K$ is given, as a user-defined variable.

In the remainder of the paper, we omit $^*$ in the notation of $L^*$ and $h^*$ for readability purposes. Although we will keep the example of skin tone, our methods apply to any continuous sensitive attribute.

\subsection{Inter-group variability as criteria for a fairness based segmentation}

We now focus on dividing $\mathcal{L}$ into $K$ groups $\mathcal{P}_k$ representing different levels of fairness, given a classification problem.
Contrary to predefined groups, these groups should be based on the level of discrimination introduced by either the ground truth $Y$ or the predicted variable $\hat{Y}$. 

Formally, for a partition $\mathcal{P}$ we define $S^\mathcal{P}$ to be a multinomial random variable associating the observations of $L$ to their group $\mathcal{P}_k$:
\begin{equation}
\label{eq:s_p_def}
    S^\mathcal{P} = k \iff L \in \mathcal{P}_k, \text{ where } k=1,\dots,K,
\end{equation}
which is commonly called the group membership. Note that while it is generally a provided variable in the dataset, it depends here on the partition $\mathcal{P}$ and is thus not fixed given an observation.
Then, we define $\Phi(S^\mathcal{P})$ as a real function of this multinomial random variable $S^\mathcal{P}$. $\Phi$ is an extension of a fairness measure $m$ computed on each group $\mathcal{P}_k$ and takes $K$ values $\Phi(k)$, $k=1,\dots,K$:
\begin{equation}
\Phi(k) = m(\mathcal{D}, \mathcal{P}_k) \in \mathbb{R}.
\label{eq:measure}
\end{equation}

For instance, a typical measure $m$ is the disparate impact, computed for samples in group $\mathcal{P}_k$ versus all other samples: 
\begin{equation}
    m(\mathcal{D}, \mathcal{P}_k) = \mathbb{P}(Y=1|L\in \mathcal{P}_k) - \mathbb{P}(Y=1|L \notin \mathcal{P}_k).
\end{equation}

We propose to find the partition $\mathcal{P}$ of the continuous sensitive attribute $L$, such that its components $\mathcal{P}_k$ have the most different levels of discrimination according to the fairness measured by $\Phi$. In other terms, such partition is the most expressive one in terms of the measure of fairness $\Phi$. Our objective is thus to find the partition $\mathcal{P}$ 
over all possible partitions that maximizes:
\begin{equation}
\label{eq:optimal_partition}
    \argmax_{\mathcal{P}} \Var\Big(\Phi \big(S^\mathcal{P} \big)\Big).
\end{equation}
where the variance is calculated over the observations of $L$ in the dataset. The variance of $\Phi(S^\mathcal{P})$ can thus be rewritten as:

\begin{equation}
    \begin{gathered}
    \Var\Big(\Phi \big(S^\mathcal{P} \big)\Big) = 
        \Var_{L \sim \mathcal{D}}\Big(\Phi \big( S^\mathcal{P} \big)\Big) = \\ =
         \int_{l \in \mathcal{L}} \Big(\Phi\big( S^\mathcal{P}\big) - \E \big[  \Phi \big( S^\mathcal{P}\big) \big]   \Big)^2 p_L(l) dl =
             \end{gathered}
             \label{eq:var_on_samples}
\end{equation}
\begin{equation}
    \begin{gathered}
 = \sum_{k=1}^K \int_{l \in \mathcal{P}_k}
        \Big(\Phi\big( k\big) - \E \big[  \Phi \big( S^\mathcal{P}\big) \big]   \Big)^2 p_L(l) dl = \\
          =
         \sum_{k=1}^{K} \mathbb{P}\big(S^\mathcal{P}=k\big)  \Big(\Phi \big(k\big) - \E \Big[  \Phi \big(S^\mathcal{P}\big) \Big] \Big)^2,
             \label{eq:var_on_clusters}
    \end{gathered}
\end{equation}
where $p_L(l)$ is the probability density function of $L$ and 
\begin{equation}
    \begin{gathered}
        \E \Big[  \Phi \big( S^\mathcal{P}\big) \Big] =
        \int_{l \in \mathcal{L}} \Phi\big( S^\mathcal{P}\big) p_L(l) dl=
        \sum_{k=1}^{K} \mathbb{P}\big(S^\mathcal{P} = k\big) \Phi \big(k \big). 
    \end{gathered}
\end{equation}

The interpretation behind this maximization is two-fold. First, from the expression over $\mathcal{D}$ of Equation \eqref{eq:var_on_samples}, we argue that the exact partition maximizes the variability of fairness outcomes across observations, so that it obtains on average the most "unfair" measures for each person of $\mathcal{D}$. Indeed, the variance in question captures the mean deviation between each observation’s group fairness measure \(\Phi\bigl(S^\mathcal{P}\bigr)\) and the population-wide expected fairness value. Second, when expressed as a sum over the groups (Equation \eqref{eq:var_on_clusters}), this same variance can be viewed as the inter-group variance---weighted by each group’s probability \(\mathbb{P}\bigl(S^\mathcal{P}=k\bigr)\). Hence, maximizing it enhances the “expressivity” of our partition by increasing the variability of fairness measures among groups.

\subsection{One-group-versus-all as fairness measure}

We propose to choose the fairness measure $m(\mathcal{D}, \mathcal{P}_k)$ (Equation \eqref{eq:measure}) based on the following.
A model $g$ is considered fair w.r.t. the sensitive attribute $S^\mathcal{P}$ if the model's output does not depend on the value of the sensitive attribute. This is usually modeled as the expected output $\hat{Y}$ being independent of the sensitive attribute $S^\mathcal{P}$:
\begin{equation}
\label{eq:fainess_sp}
    \E [\hat{Y}] = \E [\hat{Y} | S^\mathcal{P} = k], \forall k=1,\dots,K.
\end{equation}

Hence, we choose as fairness measure: 
\begin{equation}
\label{eq:choosen_phi}
\Phi(k) = m(\mathcal{D}, \mathcal{P}_k)=\mathbb{P}(Y=1 | S^\mathcal{P}=k) - \mathbb{P}(Y = 1).
\end{equation}
This measure is a natural extension of the well-known disparate impact for two classes (see Section \ref{sec:di}) and is related to the definition of subgroup fairness defined in \cite{kearns2018preventing}. 
Under this choice, Equation \eqref{eq:optimal_partition} can be rewritten in the following way
\begin{equation}
\label{eq:variance_di_1d}
\begin{gathered}
    \Var\Big(\Phi \big(S^\mathcal{P} \big)\Big)=
        \Var\Big(\mathbb{P}(Y=1 | S^\mathcal{P}) - \mathbb{P}(Y = 1)\Big) = \\ = \sum_{k=1}^{K} \mathbb{P}\big(L \in \mathcal{P}_k\big)  \Big(\mathbb{P}(Y = 1 | S^\mathcal{P} = k) - \mathbb{P}(Y = 1) \Big)^2. 
\end{gathered}
\end{equation}

Note that $\E \Big[ \Phi(S^\mathcal{P}) \Big] = \E \Big[ \mathbb{P}(Y=1 | S^\mathcal{P}) - \mathbb{P}(Y = 1) \Big]=0$.

Therefore, maximizing the variance defined in Equation \eqref{eq:variance_di_1d} when $Y$ is binary is equivalent to maximizing the sum of squared distances between $ \E [Y]$ and $\E [Y | S^\mathcal{P} = k]$ for each group $\mathcal{P}_k$. Each of the terms is multiplied by the "weight" of the group $\mathcal{P}_k$, which takes into account the impact of this group on the population. We see that with this fairness measure, the optimal partition $\mathcal{P}$ has the groups with the worst-case level of discrimination and with the major impact, or equivalently the most discriminating partition. 
In the following section, we provide a formal expression illustrating this in the case $K=2$ .

\subsection{Relation with Disparate Impact in the binary case}
\label{sec:di}
For $K=2$, we establish the relation between the $\Phi(k)$ chosen in the previous section, and the well-known Disparate Impact. The proof is in Appendix \ref{appendix:proof_di}. In the proposition below, we shift $S^\mathcal{P}$ values (Equation \eqref{eq:s_p_def}) to $0$ and $1$ to align with usual binary notations.

\begin{proposition} \label{prop:findingtheDI}
    Assume that $S^\mathcal{P}$ is a binary sensitive random variable and $\Phi(S^\mathcal{P})= \mathbb{P}(Y=1 | S^\mathcal{P}) - \mathbb{P}(Y = 1)$. Then, the partition $\mathcal{P}$ maximizing $\Var\Big(\Phi \big(S^\mathcal{P} \big)\Big)$ consists of the least and the most privileged groups with the maximal impact, i.e.
    \begin{equation}
        \Var\Big(\Phi(S^\mathcal{P})\Big) =  \pi(1-\pi)DI^2,
    \end{equation}
    where $\pi = \mathbb{P}(S^\mathcal{P}=1)$ and $DI = \mathbb{P}(Y = 1 | S^\mathcal{P}=1) - \mathbb{P}(Y = 1 | S^\mathcal{P}=0)$.
\end{proposition}

The maximized quantity $\pi(1-\pi)DI^2$ is made of two terms. First, the usual DI which promotes a separation into groups exhibiting the most different behaviors, and thus promoting unfairness. Second, the impact of the weights $\pi (1-\pi)$ enforces the probability of occurrence of the two groups to be as similar as possible, with a maximum achieved for this part at $\pi=1/2$. This provides another insight of the criterion we promote. We do not want to select a too small (or too large) group w.r.t. the other, exhibiting a particular behavior due to its size difference. Our objective is rather to identify two kind of behaviors which are intrinsically different without relying on the sampling properties of the individuals. This is coherent with the choice of so-called balance error groups when providing a mathematically sound definition for bias in the data as shown in \cite{gordaliza2019obtaining}. 

\section{Fairness aware segmentation algorithms}

In the following section, we provide methods to find the partition from Equation \eqref{eq:optimal_partition} for one- and two-dimensional sensitive variable $L \in \mathcal{L} \subset \mathbb{R}^p$, $p=1,2$. We consider partitions $\mathcal{P}$ as follows:
\begin{itemize}
    \item In the one-dimensional case, we consider $\mathcal{P}_k = [l_{k-1},l_k]$ to be segments, with $l_{k-1} < l_k$.
    \item In the two-dimensional case, we consider $\mathcal{P}_k$ to be rectangles, i.e. of the shape $[l,l'] \times [h,h']$.
\end{itemize}

\subsection{Leveraging K-Means in the case of monotonic fairness in 1D}\label{subsec:k-means}

Finding the exact solution to Equation \eqref{eq:optimal_partition} is not feasible. Hence an alternative is given by  clustering algorithms \cite{ezugwu2022comprehensive, xu2005survey}, which naturally align with a segmentation objective. Since we aim at maximizing the distance between different groups in terms of the fairness measure $\Phi$, we naturally consider applying the usual K-Means algorithm \cite{macqueen1967some} on a set of points $\psi_j \in \mathbb{R}$ whose values will aim at \textit{reflecting} this fairness. This procedure is detailed hereafter.

\begin{figure}[h]
\centering
    \includegraphics[width=\linewidth]{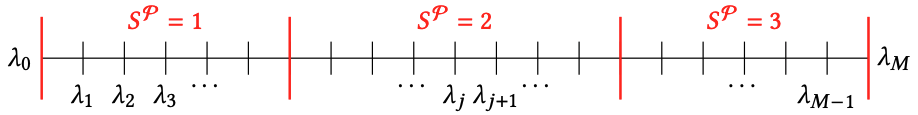}
    \caption{Partition of $L \in [\lambda_{0},\lambda_{M}]$ into intervals corresponding to $S^\mathcal{P}=1,2,3$.}
    \label{fig:discretization_k-means}
\end{figure}

Let us consider $M \in  \mathbb{N}$ and a grid $\Lambda = \{\lambda_0, \dots, \lambda_M \}$ on $\mathcal{L}$, where $\lambda_j = \lambda_{0} + j\delta$, $\delta = \frac{\lambda_M - \lambda_0}{M}$, $j = 0, \dots, M$. We compute on the $M$ intervals the  $\psi_j= \mathbb{P}(Y = 1 | L \in [\lambda_{j-1}, \lambda_j]) - \mathbb{P}(Y=1)$, $j=1,...,M$, as shown in Figure \ref{fig:discretization_k-means}. In this case, applying K-means on the $\psi_j$ will compute a partition $\mathcal{P}$ that minimizes
\begin{equation}
    \argmin_{\mathcal{P}} \sum_{k=1}^K \sum_{j=1}^M (\psi_j - \mu_k)^2 \mathds{1}_{\{\psi_j \in \mathcal{P}_k\}},
\end{equation}
where $\mu_k = \frac{\sum_{j=1}^M \psi_j \mathds{1}_{\{\psi_j \in \mathcal{P}_k\}}}{\sum_{j=1}^M \mathds{1}_{\{\psi_j \in \mathcal{P}_k\}}}$ is the centroid of the $k$-th group.

\vspace{1.0em}

By applying K-Means algorithm, the $k$-th cluster  will correspond to $\mathcal{P}_k$. The $\psi_j$ values will be homogeneous within each cluster, while being as distinct as possible between clusters.
The K-Means minimization is indeed equivalent to maximizing the sum of $(\mu_k - \mathbb{E}(\mu_k))^2$ over clusters, which relates to the maximization problem given by Equation \eqref{eq:optimal_partition}. One notes that this is an heuristic approach and the obtained $\mathcal{P}$ is not the exact solution of Equation \eqref{eq:optimal_partition}. Indeed, while groups $\mathcal{P}_k$ will be homogeneous in terms of fairness express by $\psi_j$, the exact $\Phi$ (Equation \ref{eq:measure}) is not computed on the groups $\mathcal{P}_k$, and in particular $\Phi(k)\neq \mu_k$.

Furthermore, in general, K-Means do not guarantee that groups will be connected in the $\mathcal{L}$ space. 
The underlying distance used in $L$ dimension between $\lambda_{j_1}$ and $\lambda_{j_2}$ is indeed $|\psi_{j_1}-\psi_{j_2}|$, which can take low or null values for far $\lambda_{j_1},\lambda_{j_2} \in \mathcal{L}$.
We now define the property of monotonic fairness, that will guarantee that each groups from this method is connected.
\begin{definition}
    \label{def:monotonic_fairness}
    We say that the fairness is monotonic w.r.t. to the continuous sensitive variable $L \in \mathcal{L} \subset \mathbb{R}$, if the distribution function $\mathbb{P}(Y = 1 | L = l)$ is monotonic w.r.t. $l \in \mathcal{L}$. Note that monotonicity w.r.t. $L$ is equivalent to monotonicity w.r.t. $S^{\mathcal{P}}$ $\forall \mathcal{P}$.
\end{definition}

\begin{proposition}
\label{prop:k-means_joint_clusters}
    Assume that $\mathbb{P}(Y = 1 | L )$ is monotonic w.r.t. $L \in \mathcal{L} \subset \mathbb{R}$. Then $\mathbb{P}(Y = 1 | L) - \mathbb{P}(Y=1)$ is also monotonic w.r.t. $L$, and the partition $\mathcal{P}$ found by the K-Means algorithm always consists of segments of $\mathcal{L}$.
\end{proposition}

According to Proposition \ref{prop:k-means_joint_clusters}, the assumption of monotonicity of fairness is crucial, since if it is not fulfilled, K-Means can produce groups that are disconnected (see Figure \ref{fig:di_disjoint_clusters} in Appendix \ref{appendix_di_disjoint_clusters} for details).
This notion of monotonic fairness can well occur in practice, for example in skin tone, as we'll illustrate in the following sections.

Note that $Y$ may be replaced by the model output $\hat{Y}$ in Definition \ref{def:monotonic_fairness} and Proposition \ref{prop:k-means_joint_clusters}, to analyze the model forecasts' discrimination.

\subsection{FairGroups: exhaustive search with optimal precalculation of fairness metric $\Phi$}
\label{subsec:exhaustive_search}
We now look for the solution of Equation \eqref{eq:optimal_partition} in the general case, without any assumption on function $P(Y=1 | L = l)$. We again consider a grid $\Lambda$ on $L$ with $M$ intervals. To find the partition $\mathcal{P}$ for a given $K$, we perform an exhaustive search over all possible combinations of intervals on the grid $[\lambda_{j_1}, \lambda_{j_2}]$, $j_1 < j_2$. To speed up the exhaustive search, we propose Algorithms \ref{alg:iteratively_count} and \ref{alg:calculate_triangle} (Appendix \ref{appendix:exhaustive_search}) to efficiently precalculates $\Psi_{j_1, j_2}:= \mathbb{P}(Y = 1 | L \in [\lambda_{j_1}, \lambda_{j_2}]) - \mathbb{P}(Y=1)$, $\forall j_1 < j_2$ with a complexity $O(M^2)$ using a dynamic programming strategy.

For this, we define an upper triangular matrix $U_{\Psi_\Lambda} \in \mathbb{R}^{M \times M}$ where we store all the precalculated values as follows
\begin{equation*}
(U_{\Psi_\Lambda})_{i, j} =
    \begin{cases}
      \Psi_{i-1, j}, & \text{ for } i \leq j, \text{ where } i, j = 1, \dots, M-1,\\
      0, & \text{ for } i > j.
    \end{cases}
\end{equation*}
Each value $\Psi_{j_1, j_2}$ of matrix $U_{\Psi_\Lambda}$ is estimated using the ratio between $\sum_{j_1 < j_2}  \sum_{\mathcal{D}} \mathds{1}_{\{L \in [\lambda_{j_1}, \lambda_{j_2}]\}} \mathds{1}_{\{Y=1\}}$ and $\sum_{j_1 < j_2}  \sum_{\mathcal{D}} \mathds{1}_{\{L \in [\lambda_{j_1}, \lambda_{j_2}]\}}$, that both can be computed by recurrence on $\lambda_{j_1}$ and $\lambda_{j_2}$ using Algorithm \ref{alg:iteratively_count}.
Matrix $U_{\Psi_\Lambda}$ is then efficiently precomputed with Algorithm \ref{alg:calculate_triangle} and is subsequently used for exhaustive search. This eliminates the need to recalculate the values of $\Psi_{j_1, j_2}$, when evaluating potential combination of intervals, thereby significantly speeding up the computational process.
One notes that given the grid $\Lambda$, the $\psi_j$ defined in the previous section are the diagonal of $U_{\Psi_\Lambda}$, since $\psi_j=\Psi_{j-1,j}$

Given this matrix $U_{\Psi_\Lambda}$, we find the partition $\mathcal{P}$ of segments $\mathcal{P}_k = [\lambda_{j_{k-1}}, \lambda_{j_{k}}]$ $(j_{k-1} < j_{k})$ of Equation \eqref{eq:optimal_partition} as:
\begin{equation}
    \begin{aligned}
    \argmax_{\substack{0<j_1<j_2<\dots\\\dots<j_{K-1}<M}} \quad \sum_{k=1}^K \mathbb{P}\big(L \in [\lambda_{j_{k-1}}, \lambda_{j_k}]\big) \left(
\Psi_{j_{k-1},j_k} - \E \left[  \Phi\big(S^{\mathcal{P}}\big) \right]
\right)^2,
    \end{aligned}
\end{equation}
where $\E [\Phi \big(S^\mathcal{P}\big)]=\sum_{k=1}^K \mathbb{P}(L\! \in \! [\lambda_{j_{k-1}}, \lambda_{j_{k}}]) 
\Psi_{j_{k-1},j_{k}} $. The $\mathbb{P}(L \in [\lambda_{j_{k-1}}, \lambda_{j_{k}}])$ values are precalculated using Algorithm 1. 
The solution is then exact, up to the approximation of the partitions search space induced by the grid $\Lambda$.

\section{Experiments}
The code for using our methods and reproducing these experiments is in open access \footnote{\texttt{https://github.com/artefactory/fairness\_metrics}}. 

\begin{table*}[htbp]
\begin{center}
\caption{Comparison of partitions of $L$ for our synthetic datasets based on $\Phi(S^\mathcal{P}) = \mathbb{P}(Y = 1 | S^\mathcal{P}) - \mathbb{P}(Y = 1)$, $K = 5$.}
\label{table:metrics_uniform_gaussian}
\begin{tabular}{|c|cc|cc|}
\hline
Data distribution for computing $\mathcal{P}$
& \multicolumn{2}{c|}{$L \sim \mathcal{U}(0,100)$} & \multicolumn{2}{c|}{$L \sim \mathcal{N}(50, 20)$ truncated on $[0,100]$}
\\ \hline
Metric & \multicolumn{1}{c|}{$\Var\Big(\Phi\big(S^{\mathcal{P}}\big)\Big)$} & \begin{tabular}[c]{@{}c@{}}Rand Index\\ (with ground truth)\end{tabular} & \multicolumn{1}{c|}{$\Var\Big(\Phi\big(S^{\mathcal{P}}\big)\Big)$} & \begin{tabular}[c]{@{}c@{}}Rand Index\\ (with ground truth)\end{tabular} 
\\ \hline
$\mathcal{P}$: Ground Truth & \multicolumn{1}{c|}{\textbf{0.068}} & 1 & \multicolumn{1}{c|}{\textbf{0.032}} & 1 
\\ \hline
$\mathcal{P}$: K-Means & \multicolumn{1}{c|}{0.067}  & 0.972 & \multicolumn{1}{c|}{0.025} & 0.804 
\\ \hline
$\mathcal{P}$: FairGroups & \multicolumn{1}{c|}{\textbf{0.068}} & 0.99 & \multicolumn{1}{c|}{\textbf{0.032}} & 0.97
\\ \hline
\end{tabular}
\end{center}
\end{table*}

\begin{table*}[htpb]
\centering
\caption{Comparison of partitions of $IT\!A$ based on $\Phi(S^\mathcal{P}) = \mathbb{P}(\hat{Y} = 1 | S^\mathcal{P}) - \mathbb{P}(\hat{Y} = 1)$, $K = 6$.}
\label{table:ita_metrics_k=6}
\begin{tabular}{|c|c|c|c|c|}
\hline
             Data for computing $\mathcal{P}$
             & \multicolumn{2}{c|}{CelebA} & \multicolumn{1}{c|}{FFHQ} 
             & CelebA \& FFHQ
\\ \hline
\begin{tabular}[c]{@{}c@{}}Metric \\ \end{tabular} & \begin{tabular}[c]{@{}c@{}}$\Var\Big(\Phi\big(S^{\mathcal{P}}\big)\Big)$\\on CelebA\end{tabular} & \begin{tabular}[c]{@{}c@{}}$\Var\Big(\Phi\big(S^{\mathcal{P}}\big)\Big)$\\on FFHQ\end{tabular} & \begin{tabular}[c]{@{}c@{}}$\Var\Big(\Phi\big(S^{\mathcal{P}}\big)\Big)$\\on FFHQ\end{tabular} & \begin{tabular}[c]{@{}c@{}}Rand Index\\ between both partitions,\\ on FFHQ\end{tabular}
\\ \hline
$\mathcal{P}$: Fitzpatrick & 0.078  & 0.079 & 0.079 & 1
\\ \hline
$\mathcal{P}$: K-Means & 0.089 & 0.087 & 0.088 & 0.869                 
\\ \hline
$\mathcal{P}$: FairGroups & \textbf{0.092} & \textbf{0.09} & \textbf{0.093} & 0.921
\\ \hline
\end{tabular}
\end{table*}

\subsection{Synthetic datasets}
We first evaluate our proposed methods on a series of synthetic data.
To do so, we suppose that the discrimination is expressed in the dataset in terms of Statistical Parity, i.e. we have distinct intervals with different probability of success $\mathbb{P}(Y = 1 | L)$.
We also choose to work in the scenario of monotonic fairness (Proposition \ref{prop:k-means_joint_clusters}).

\begin{figure}[h]
\centering
     \begin{subfigure}{0.47\linewidth}
        \centering
        \includegraphics[width=\linewidth]{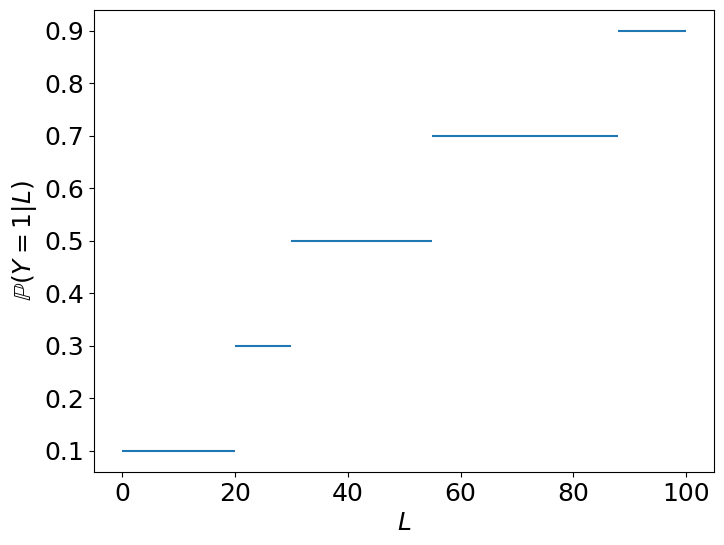}  
        \label{fig:proba_synthetic_data}
    \end{subfigure}
    \hfill
    \begin{subfigure}{0.49\linewidth}
        \centering
        \includegraphics[width=\linewidth]{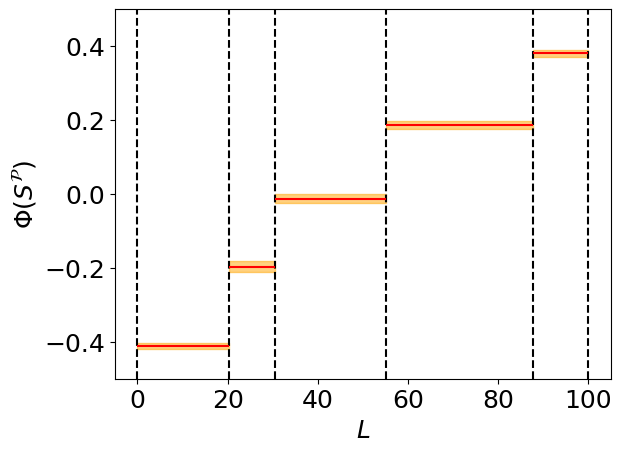}  
        \label{fig:ground_truth_uniform}
    \end{subfigure}
\caption{Partitions on synthetic dataset with uniform distribution of $L$: $\mathbb{P}(Y = 1 | L)$ on the left and FairGroups partition on the right.}
\vspace{1.5em}
\label{fig:partitions_synthetic_data_uniform}
\end{figure}

We suppose that $L \sim \mathcal{U}(0,100)$ and corresponding $Y \sim  Bernoulli(p(L))$, where $p(L) = 0.1 \times\mathds{1}_{\{L \leq 20\}} + 0.3 \times\mathds{1}_{\{20 < L \leq 30\}} + 0.5\times\mathds{1}_{\{30 < L \leq 55\}} + 0.7\times\mathds{1}_{\{55 < L \leq 88\}} + 0.9\times\mathds{1}_{\{88 < L \leq 100\}}$ (see Figure \ref{fig:partitions_synthetic_data_uniform}). First, we sample $N=50000$ pairs of observations $(L, Y)$ from this distribution $\mathcal{D}$. Second, to illustrate a non-uniform scenario, we additionally sample a dataset in the case $L$ follows a truncated normal distribution on $[0,100]$ from $\sim \mathcal{N}(50,20)$. The two empirical distributions are shown in Figure \ref{fig:synthetic_data} (Appendix \ref{appendix:synthetic_data_gaussian}).


On both datasets, we apply the K-Means and the FairGroups methods to find the partition of $L$, as shown in Figure \ref{fig:partitions_synthetic_data_uniform}. We compute $\Phi(S^\mathcal{P})$ values for each group $\mathcal{P}_k$ in y-axis, as well as their asymptotic confidence intervals using the delta method following the work \cite{besse2018confidence} (as red transparent areas in Figures \ref{fig:partitions_synthetic_data_uniform} and \ref{fig:partitions_synthetic_data_gaussian} in Appendix \ref{appendix:synthetic_data_gaussian}).
To measure how close each found partition is to the ground truth partition, we measure the pairwise Rand index \cite{rand1971objective} with the ground truth partition, a common measure of the similarity between two data clusterings. 
One notes that the ground truth partition is independent of the $L$ distribution and is thus the same for both synthetic datasets.
Table \ref{table:metrics_uniform_gaussian} shows results alongside the variance $\Var\big(\Phi(S^{\mathcal{P}})\big)$ values.

From Figure \ref{fig:partitions_synthetic_data_uniform} and Table \ref{table:metrics_uniform_gaussian}, we conclude that FairGroups approach managed to find the ground truth partition almost perfectly, i.e. the Rand Index is 0.99. Furthermore, we can see that FairGroups manages to find the ground truth partition of $L$ regardless of its distribution, with very high Rand Index for the non-uniform distribution. 
On the other hand, K-Means groups remains quite close to the ground truth in the case of uniform data, but is less accurate than FairGroups. For non-uniform data, we see that K-Means Rand Index is even lower.


\subsection{Data and protocol for face analysis w.r.t skin tone}
\label{sec:data_protocol}
In this section, we use the women face pictures of CelebA dataset \cite{liu2015faceattributes} as $\mathcal{D}$, with the number of observations $N=84602$. We predict skin tone $L, a, b$ values for each image in $\mathcal{D}$ using the proprietary algorithm \cite{robin2020beyond}, and derive from it $L$, $IT\!A$ or $(L, h)$ as the sensitive variable. As the target output variable $Y$, we use the attribute "Attractive". We fine-tune on 5 epochs a ResNet18 \cite{he2016deep} as model $g_\theta$, without any bias mitigation strategy, with batch size equal to 64. The prediction $\hat{Y}$ for each CelebA image is obtained after a 5-fold cross-validation, stratified on $10$ equal-sized bins of $L$.
We also consider FFHQ\footnote{\texttt{https://github.com/NVlabs/ffhq-dataset}} ($N = 26906$) 
for assessing the generalization of our partitioning algorithm, with skin tone extracted as for CelebA pictures. FFHQ does not have a ground truth $Y$ for our task, so that we predict $\hat{Y}$ as the mean of our models $g_\theta$, meaning we are looking at the discrimination of the same model. Please note that the sensitive attribute values being estimated by a proprietary algorithm, they cannot be publicly shared.

The "Attractive" target on CelebA dataset represents an interesting case of monotonic fairness (Definition \ref{def:monotonic_fairness}), as shown in Figure \ref{fig:partitions_celeba}. We indeed observe in the initial data as well as in the model's predictions that, on average, the darker a person's skin is, the more discriminated they are. Please note that this target is known to represent a biased annotation \cite{risser2022tackling}, and was chosen purely for illustrative purpose.

\subsection{Partitions of skin tone in 1D}
\label{section:partitions_ita}
\begin{figure}[h]
\centering
    \begin{subfigure}{\linewidth}
        \centering
        \includegraphics[width=0.5\linewidth]{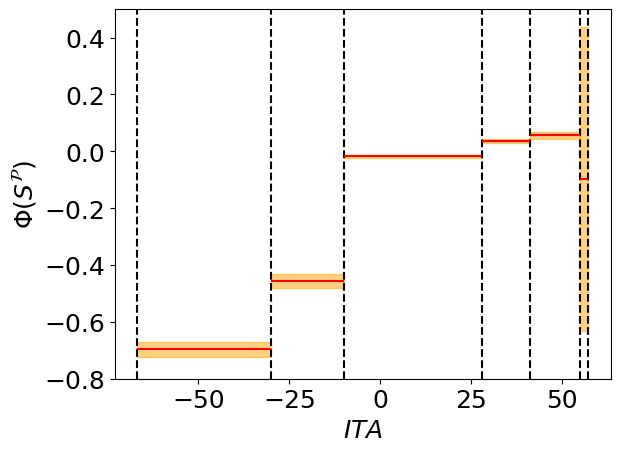}  
    \end{subfigure}
    \begin{subfigure}{0.49\linewidth}
        \centering
        \includegraphics[width=\linewidth]{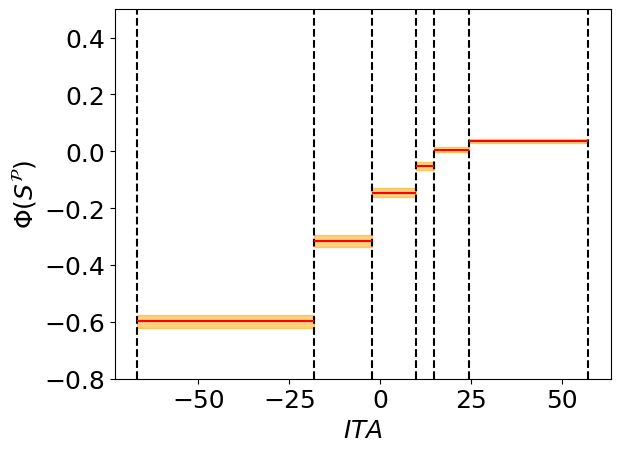}  
    \end{subfigure}
    \hfill    
    \begin{subfigure}{0.49\linewidth}
        \centering
        \includegraphics[width=\linewidth]{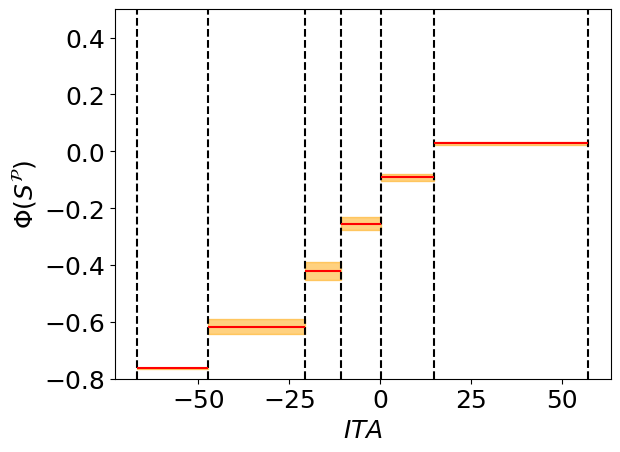}  
    \end{subfigure}
\caption{Partitions of $IT\!A$ using different methods. Fitzpatrick (default) partition of $IT\!A$ (top). FairGroups partition of $IT\!A$ (bottom left). K-Means partition of $IT\!A$ (bottom right).}
\vspace{1.5em}
\label{fig:partitions_ita_celeba}
\end{figure}

We estimated partitions of skin tone given by $IT\!A$ and compare them to the default partition of $IT\!A$ with Fitzpatrick scale in 6 groups. Figure \ref{fig:partitions_ita_celeba} and Table \ref{table:ita_metrics_k=6} show the results comparing the different groupings.
Both FairGroups and K-Means partitions identify groups with varying levels of discrimination. The Fitzpatrick scale, as a default partition, does not consider the underlying data distribution, so that the confidence intervals for certain groups are relatively broad, so that it is difficult to draw conclusions regarding fairness between groups.

We also consider the partitions of FFHQ and  compare them with the partitions of CelebA applied to FFHQ, as done in the previous subsection. From Table \ref{table:ita_metrics_k=6}, we observe that both FairGroups and K-Means partitions remain stable when the data distribution changes. This suggests that the partitions effectively reveal how discrimination manifests within the sensitive attribute space.

\begin{table*}[htpb]
\centering
\caption{Comparison of partitions of $(L,h)$ in 2D based on $\Phi(S^\mathcal{P}) = \mathbb{P}(\hat{Y} = 1 | S^\mathcal{P}) - \mathbb{P}(\hat{Y} = 1)$, $K = 4$.}
\label{table:l_h_metrics_k=4}
\begin{tabular}{|c|c|c|c|c|}
\hline
             Data for computing $\mathcal{P}$
             & \multicolumn{2}{c|}{CelebA} & \multicolumn{1}{c|}{FFHQ} 
             & CelebA \& FFHQ
\\ \hline
\begin{tabular}[c]{@{}c@{}}Metric \\ \end{tabular} & \begin{tabular}[c]{@{}c@{}}$\Var\Big(\Phi\big(S^{\mathcal{P}}\big)\Big)$\\on CelebA\end{tabular} & \begin{tabular}[c]{@{}c@{}}$\Var\Big(\Phi\big(S^{\mathcal{P}}\big)\Big)$\\on FFHQ\end{tabular} & \begin{tabular}[c]{@{}c@{}}$\Var\Big(\Phi\big(S^{\mathcal{P}}\big)\Big)$\\on FFHQ\end{tabular} & \begin{tabular}[c]{@{}c@{}}Rand Index\\ between both partitions,\\ on FFHQ\end{tabular}
\\ \hline
$\mathcal{P}$: Default & 0.054 & 0.082 & 0.082 & 1                                       
\\ \hline
$\mathcal{P}$: FairGroups & \textbf{0.084} & \textbf{0.102}& \textbf{0.105} & 0.883 
\\ \hline
\end{tabular}
\end{table*}

\subsection{Amplification of skin color bias by the model}
The proposed data-driven segmentation can also uncover whether discrimination is exacerbated when a model is trained on the target task. This is indeed the case if the differences between  $\Phi(S^\mathcal{P})$ values for the ground truth $Y$ and the model's output $\hat{Y}$ are significant.

\begin{figure}[h]
\centering
    \begin{subfigure}{0.46\linewidth}
        \centering
        \includegraphics[width=\linewidth]{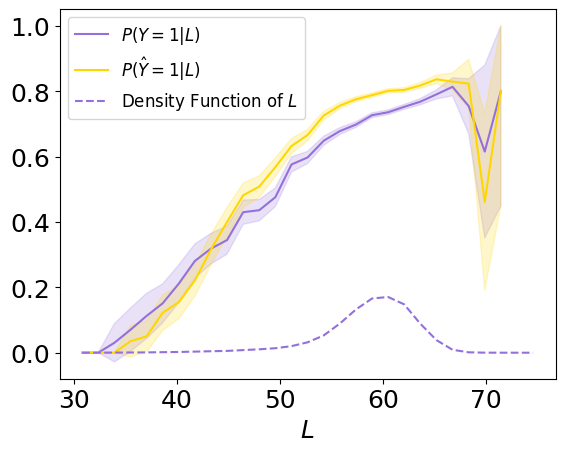}  
        \label{fig:y_proba_cond_s}
    \end{subfigure}
    \hfill
    \begin{subfigure}{0.49\linewidth}
        \centering
        \includegraphics[width=\linewidth]{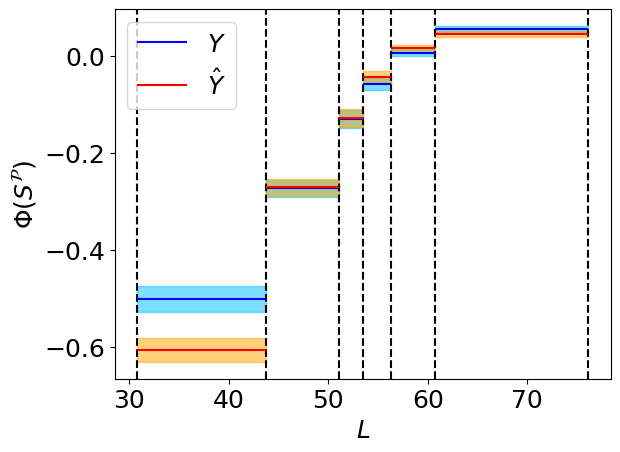}  
        \label{fig:celeba_partition_y_y_hat}
    \end{subfigure}
\caption{Bias amplification for dark skin tones. Conditional distribution of "Attractive" ($Y/\hat{Y}$) w.r.t. $L$ (left). FairGroups partition of $L$ based on $\Phi(S^\mathcal{P})$ and $Y$ (right).
\label{fig:partitions_celeba}
}
\vspace{1.5em}
\end{figure}

To demonstrate that the model amplifies discrimination on CelebA, we first compute FairGroups partition $\mathcal{P}$ by using $Y$ as a target variable. We then use this partition to recalculate the values of $\Phi(S^\mathcal{P})$ for $\hat{Y}$ as a target (Equation \ref{eq:choosen_phi}). Figure \ref{fig:partitions_celeba} illustrates that the discrimination is significantly amplified for the darkest skin tones ($L < 43$) in the CelebA dataset, since the fairness measure $\Phi(S^\mathcal{P})$ calculated on the predicted $\hat{Y}$ is significantly lower than when computed on the ground truth $Y$.

\subsection{Extension to partitioning skin tone in 2D}
\begin{figure}[h]
\centering
    \begin{subfigure}{0.47\linewidth}
        \centering
        \includegraphics[width=\linewidth]{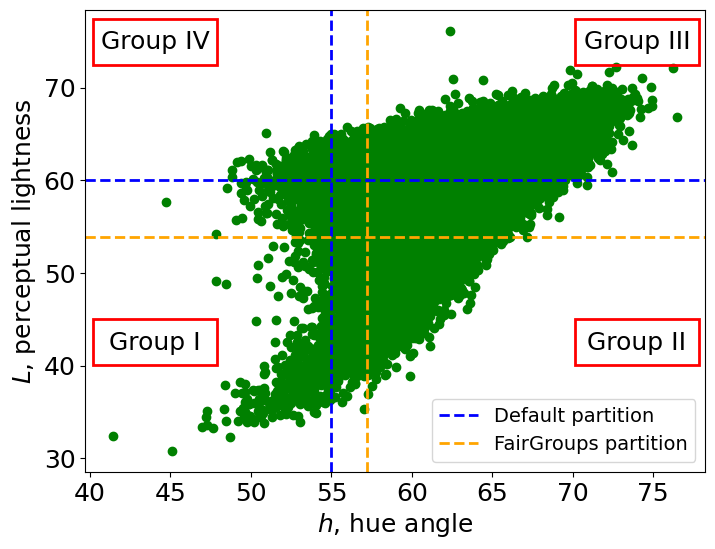}  
        \label{fig:celeba_l_h_partitions}
    \end{subfigure}
    \hfill
    \begin{subfigure}{0.49\linewidth}
        \centering
        \includegraphics[width=\linewidth]{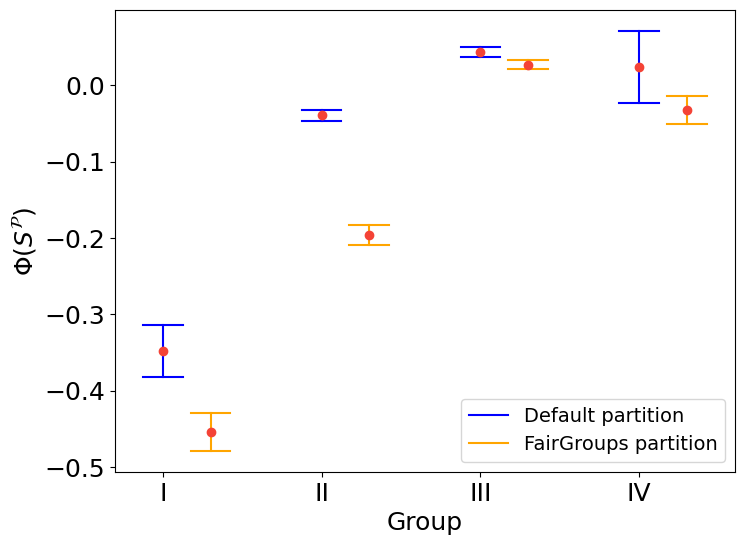}
        \label{fig:celeba_l_h_ci}
    \end{subfigure}
\caption{Default and FairGroups Partition of $(L,h)$ into 4 groups on CelebA (left), and their $\Phi(S^{\mathcal{P}})$ with Confidence intervals (right).}
\label{fig:partitions_celeba_2D}
\end{figure}

We now compare different partitions in the two-dimensional space $(L,h)$. In the literature, the default partition into 4 groups with the thresholds $L \leq 60$ and $h \leq 55$ is used to conclude about fairness \cite{thong2023beyond}. In Figure \ref{fig:partitions_celeba_2D}, we compare the partition $\mathcal{P}$ computed by FairGroups with this default partition in the left, and  the confidence intervals for $\Phi(S^\mathcal{P})$ on the right. 
We see that FairGroups partition identifies two discriminated groups (Groups I and II), while the default partition captures only one (Group I). Indeed, according to the default partition, Group II has a value $\Phi(S^\mathcal{P})$ close to 0, suggesting that this group is not discriminated against. In contrast, the same group identified by FairGroups, located in the same quadrant, has a $\Phi(S^\mathcal{P})$ close to -0.2, indicating discrimination. Additionally, FairGroups partition finds a Group I facing a higher level of discrimination than the equivalent in default segmentation. Moreover, we note that the confidence intervals for $\Phi(S^\mathcal{P})$ in FairGroups partition are narrower compared to those of the default groups.

In Table \ref{table:l_h_metrics_k=4}, we compute the $\mathbb{V}ar\big(\Phi(\mathcal{S}^\mathcal{P})\big)$ values for the two partitions on CelebA and FFHQ, and observe significantly higher values for the one from FairGroups.
We also apply the CelebA partition to FFHQ and demonstrate that it closely aligns with the skin tone partition of FFHQ (see Table \ref{table:l_h_metrics_k=4}), thus revealing how discrimination manifests in the skin tone space.

\section{Application to model debiasing}

In this section, we leverage the proposed partitions to mitigate bias at inference time through a post-processing strategy.
Specifically, we adopt the optimal transport-based approach introduced in \cite{gouic2020projection}, which maps the conditional distribution of model scores for each group to a fair target distribution. 
This approach is supported by a theoretical result related to Statistical Parity (SP): the minimum classification error achievable under SP constraints corresponds to the solution of a Wasserstein barycenter problem \cite{gouic2020projection}. In this framework, the barycenter serves as a common target distribution to which group-conditional score distributions are aligned, ensuring SP across groups while minimizing the loss in predictive accuracy.

In practice, our methodology closely follows the procedure outlined in \cite{gouic2020projection} and \cite{xian2023fair}, where the predicted score distribution for a group $\mathcal{P}_k$ is adjusted via a transformation of its cumulative distribution function. Let $\eta_k(X) = \mathbb{E}(Y|X,L \in \mathcal{P}_k)$ denote the conditional expected score for group $\mathcal{P}_k$; the goal is to map this distribution to a group-specific target distribution $g_k^*$ by solving:

\begin{equation}
    \min_{\substack{\forall k, g_k \in \mathcal{G} \\ \forall k,k' \|g_k-g_{k'}\|_\infty < \alpha}} \sum_{k=1}^K \frac{\mathbb{P}(L \in \mathcal{P}_k)}{2} W_1(\eta_k, g_k),
\end{equation}
where $\alpha \in [0,1]$ is a user parameter, $W_1$ the Wasserstein-1 distance and $\mathcal{G}$ the class of scorers in $X$ space ($g_k^*$ representing $\mathbb{P}(\hat Y\!=\!1|X,L\!\in\! \mathcal{P}_k)$). We select $W_1$ as our application focuses on binary classification; please note that for regression tasks or scenarios requiring Equality of Odds, $W_2$ is more appropriate, as detailed in \cite{gouic2020projection}. The details for this post-processing is beyond the scope of this paper (see Appendix \ref{appendix:debiasing} for  experimental illustrations).

\begin{table}[h]
\centering
\caption{Comparison of model accuracy, PR-AUC and fairness (HGR) without and with post-processing applied on partitions $\mathcal{P}$ given by different grouping methods.}
\label{table:bias_mitigation_results}
\begin{tabular}{l|c|c|c|}
\cline{2-4}
\multicolumn{1}{c|}{}                             & Accuracy & PR-AUC & HGR$(\hat{Y}, IT\!A)$   \\ \hline
\multicolumn{1}{|l|}{No post-processing}          & 0.793    & 0.824             & 0.126 \\ \hline
\multicolumn{1}{|l|}{$\mathcal{P}$ : Fitzpatrick} & 0.782    & 0.815             & 0.095 \\ \hline
\multicolumn{1}{|l|}{$\mathcal{P}$ : K-Means}     & 0.783    & 0.813             & 0.065 \\ \hline
\multicolumn{1}{|l|}{$\mathcal{P}$ : FairGroups}  & 0.781    & 0.811             & \textbf{0.039} \\ \hline
\end{tabular}
\end{table}

We evaluated this post-processing on the model $g_\theta$ obtained using the protocol of Section \ref{sec:data_protocol}. We calculated PR-AUC \cite{zhu2004recall} and HGR \cite{mary2019fairness} between predictions and $IT\!A$ using the corrected $\mathbb{P}(\hat{Y}=1|X)$ predicted on the successive test sets.
As shown in Table~\ref{table:bias_mitigation_results}, post-processing with FairGroups achieves the lowest HGR score, indicating a significant reduction in dependence between the predictions and the sensitive attribute, with only a minor reduction of accuracy and PR-AUC. This appears to be the best trade-off between predictive performance and fairness, which demonstrates the effectiveness of FairGroups partition to mitigate the bias of a model.

Note that given a sample $X$, this post-processing requires to know the sensitive attribute $L$ in order to debias the predicted $\hat{Y}$, which is accessible at inference time for our industrial scenario \cite{robin2020beyond}.

\section{Conclusion}

In this paper, we proposed FairGroups, a data-driven algorithm aiming to divide a continuous sensitive variable into groups that reflects the unfairness of a given task. We observed on synthetic and real-world datasets that our method successfully captures the specificities of the target task while taking into account the group sizes, contrary to predefined groups.
FairGroups partitions obtain better evaluation metrics, and appear stable across various data distributions.

We performed extensive experiments on the case of skin tone on CelebA and FFHQ face pictures, which is a typical sensitive attribute that we extracted as a continuous physical quantity using a proprietary algorithm. 
We observed the property of monotonic fairness w.r.t. skin tone, for which K-Means grouping provides a practical and effective approximation to the exact solution. We could also apply and validate FairGroups partitions in one and two-dimensional skin tone spaces, and obtain the best trade-off between fairness and accuracy when applying on it model debiasing based on an optimal-transport post-processing.

As future work, we will study the impact of the choice of $K$ and investigate its optimal value, and improve the scalability of our algorithm in the case of multi-dimensional sensitive attribute.





\newpage
\section*{Ethics Statement}
Considering the potential subjectivity of any index based on physical appearance, our work falls into the general problem of ensuring that, in deployment, the model remains unbiased with respect to skin tone. This means that the quality of predictions should not be affected by a user’s skin tone, and thus leading to an infringement of fundamental rights as defined in Article 21 of the charter of fundamental rights of the European Union. Hence to guarantee legal conformity of the AI system, appropriate bias measurement tools are required to assess whether the model exhibits discriminatory behavior.





\bibliography{mybibfile}

\newpage
\appendix

\section{Relation with Disparate Impact in the Binary Case}\label{appendix:proof_di}

Detailed Proof of Proposition~\ref{prop:findingtheDI}.
\begin{proof}
    When $S^\mathcal{P}$ is binary, that is, there are only two groups, the variance in Equation \eqref{eq:variance_di_1d} rewrites as follows.
\begin{equation}
    \begin{gathered}
        \Var\Big(\mathbb{P}(Y=1 | S^\mathcal{P}) - \mathbb{P}(Y = 1)\Big) = \\ = \mathbb{P}(S^\mathcal{P} = 0)\Big[\mathbb{P}(Y = 1 | S^\mathcal{P}=0) - \mathbb{P}(Y = 1) \Big]^2 + \\ + \mathbb{P}(S^\mathcal{P} = 1)\Big[\mathbb{P}(Y = 1 | S^\mathcal{P}=1) - \mathbb{P}(Y = 1) \Big]^2 = \\ = \mathbb{P}(S^\mathcal{P} = 0)\Big[\mathbb{P}(Y = 1 | S^\mathcal{P}=0) - \\ - \mathbb{P}(S^\mathcal{P} = 0)\mathbb{P}(Y = 1 | S^\mathcal{P}=0) - \\ - \mathbb{P}(S^\mathcal{P} = 1)\mathbb{P}(Y = 1 | S^\mathcal{P}=1) \Big]^2 + \\ + \mathbb{P}(S^\mathcal{P} = 1)\Big[\mathbb{P}(Y = 1 | S^\mathcal{P}=1) - \\ - \mathbb{P}(S^\mathcal{P} = 0)\mathbb{P}(Y = 1 | S^\mathcal{P}=0) - \\ - \mathbb{P}(S^\mathcal{P} = 1)\mathbb{P}(Y = 1 | S^\mathcal{P}=1) \Big]^2 = \\= \mathbb{P}(S^\mathcal{P} = 0)\Big[(1 - \mathbb{P}(S^\mathcal{P} = 0))\mathbb{P}(Y = 1 | S^\mathcal{P}=0) - \\ - \mathbb{P}(S^\mathcal{P} = 1)\mathbb{P}(Y = 1 | S^\mathcal{P}=1) \Big]^2 + \\ + \mathbb{P}(S^\mathcal{P} = 1)\Big[(1 - \mathbb{P}(S^\mathcal{P} = 1))\mathbb{P}(Y = 1 | S^\mathcal{P}=1) - \\ - \mathbb{P}(S^\mathcal{P} = 0)\mathbb{P}(Y = 1 | S^\mathcal{P}=0) \Big]^2 = \\ = \mathbb{P}(S^\mathcal{P} = 0)\Big[\mathbb{P}(S^\mathcal{P} = 1)\mathbb{P}(Y = 1 | S^\mathcal{P}=0) - \\ - \mathbb{P}(S^\mathcal{P} = 1)\mathbb{P}(Y = 1 | S^\mathcal{P}=1) \Big]^2 + \\ + \mathbb{P}(S^\mathcal{P} = 1)\Big[\mathbb{P}(S^\mathcal{P} = 0)\mathbb{P}(Y = 1 | S^\mathcal{P}=1) - \\ - \mathbb{P}(S^\mathcal{P} = 0)\mathbb{P}(Y = 1 | S^\mathcal{P}=0)\Big]^2 = \\ = \mathbb{P}(S^\mathcal{P} = 0)\mathbb{P}(S^\mathcal{P} = 1)^2\Big[\mathbb{P}(Y = 1 | S^\mathcal{P}=0) - \\ - \mathbb{P}(Y = 1 | S^\mathcal{P}=1) \Big]^2 + \\ + \mathbb{P}(S^\mathcal{P} = 1)\mathbb{P}(S^\mathcal{P} = 0)^2\Big[\mathbb{P}(Y = 1 | S^\mathcal{P}=1) - \\ - \mathbb{P}(Y = 1 | S^\mathcal{P}=0)\Big]^2 = \\ = \mathbb{P}(S^\mathcal{P}=0)\mathbb{P}(S^\mathcal{P}=1)(\mathbb{P}(S^\mathcal{P}=0) + \\ + \mathbb{P}(S^\mathcal{P}=1))\Big[\mathbb{P}(Y = 1 | S^\mathcal{P}=0) - \mathbb{P}(Y = 1 | S^\mathcal{P}=1)\Big]^2 \\ = \mathbb{P}(S^\mathcal{P}=0)\mathbb{P}(S^\mathcal{P}=1)\Big[\mathbb{P}(Y = 1 | S^\mathcal{P}=0) - \\ - \mathbb{P}(Y = 1 | S^\mathcal{P}=1)\Big]^2 = \\ = \mathbb{P}(S^\mathcal{P}=0)\mathbb{P}(S^\mathcal{P}=1)DI^2 = \pi(1-\pi)DI^2.
    \end{gathered}
\end{equation}
\end{proof}

\section{K-Means in the case of non-monotonic fairness: disconnected clusters}
\label{appendix_di_disjoint_clusters}

\begin{figure}[h]
    \centering
    \includegraphics[width=0.45\textwidth]{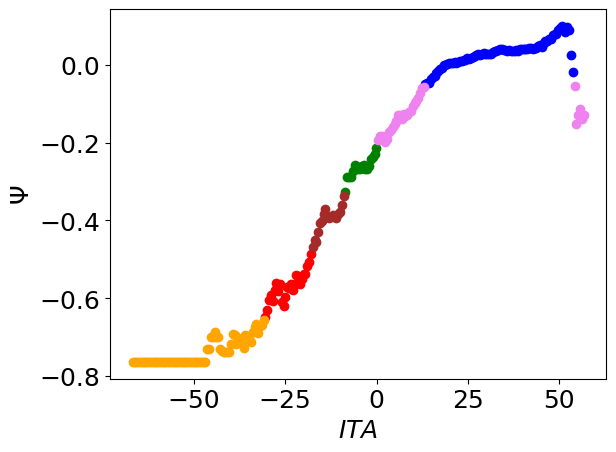}
    \caption{K-Means produces disconnected groups in the case of non-monotonic fairness.}
    \label{fig:di_disjoint_clusters}
\end{figure}

If fairness is not monotonic (Definition \ref{def:monotonic_fairness}), then K-Means may produce disconnected groups as illustrated in Figure \ref{fig:di_disjoint_clusters}. We observe that pink group is separated into two segments by blue group, since K-Means only use values of $\Psi$ for grouping.

\vspace{1.0em}

\section{Exhaustive search with optimal precalculation of fairness metric $\Phi$}\label{appendix:exhaustive_search}

We can approximate upper triangular matrix $U_{\Psi_\Lambda}$, defined in \ref{subsec:exhaustive_search}, as follows
\begin{equation}
    \begin{gathered}
    U_{\Psi_\Lambda} \approx \\ \approx
    \begin{bmatrix}
        \sum_{i=1}^N\mathds{1}_{\{Y_i = 1, L_i \in [\lambda_0, \lambda_1]\}} & \dots & \sum_{i=1}^N\mathds{1}_{\{Y_i = 1, L_i \in [\lambda_0, \lambda_{M-1}]\}} \\
        0 & \dots & \sum_{i=1}^N\mathds{1}_{\{Y_i = 1, L_i \in [\lambda_1, \lambda_{M-1}]\}} \\
        \dots & \dots & \dots \\
        0 & \dots & \sum_{i=1}^N\mathds{1}_{\{Y_i = 1, L_i \in [\lambda_{M-2}, \lambda_{M-1}]\}}
    \end{bmatrix} \odot \\ \odot
    \begin{bmatrix}
        \sum_{i=1}^N\mathds{1}_{\{L_i \in [\lambda_0, \lambda_1]\}} & \dots & \sum_{i=1}^N
        \mathds{1}_{\{L_i \in [\lambda_0, \lambda_{M-1}]\}} \\
        0 & \dots & \sum_{i=1}^N\mathds{1}_{\{L_i \in [\lambda_1, \lambda_{M-1}]\}} \\
        \dots & \dots & \dots \\
        0 & \dots & \sum_{i=1}^N\mathds{1}_{\{L_i \in [\lambda_{M-2}, \lambda_{M-1}]\}}
    \end{bmatrix}^{\circ-1}
    - \\ - \frac{\sum_{i=1}^N\mathds{1}_{\{Y_i = 1\}}}{N} \times U,
    \end{gathered}
    \label{eq:matrix_u_approx}
\end{equation}

where $\odot$ is Hadamard (element-wise) product and $^{\circ-1}$ is Hadamard inverse, U is an upper triangular matrix of size $(M-1) \times (M-1)$ with ones on the main diagonal and in all elements above it. Therefore, we have
\begin{equation}
    \begin{gathered}
        \Psi_{j_1, j_2} \approx \frac{\sum_{i=1}^N\mathds{1}_{\{Y_i = 1\}}\mathds{1}_{\{L_i \in [\lambda_{j_1-1}, \lambda_{j_2}]\}}}{\sum_{i=1}^N\mathds{1}_{\{L_i \in [\lambda_{j_1-1}, \lambda_{j_2}]\}}} - \\ - \frac{\sum_{i=1}^N\mathds{1}_{\{Y_i = 1\}}\mathds{1}_{\{L_i \in [\lambda_{j_1-1}, \lambda_{j_1}]\}}}{\sum_{i=1}^N\mathds{1}_{\{L_i \in [\lambda_{j_1-1}, \lambda_{j_1}]\}}} - \frac{\sum_{i=1}^N\mathds{1}_{\{Y_i = 1\}}}{N} = \\ = \frac{\sum_{i=1}^N\mathds{1}_{\{Y_i = 1\}}\Big(\mathds{1}_{\{L_i \in [\lambda_{j_1-1}, \lambda_{j_2}]\}} - \mathds{1}_{\{L_i \in [\lambda_{j_1-1}, \lambda_{j_1}]\}}\Big)}{\sum_{i=1}^N\Big(\mathds{1}_{\{L_i \in [\lambda_{j_1-1}, \lambda_{j_2}]\}} - \mathds{1}_{\{L_i \in [\lambda_{j_1-1}, \lambda_{j_1}]\}}\Big)} - \\ - \frac{\sum_{i=1}^N\mathds{1}_{\{Y_i = 1\}}}{N} = \\ = \frac{\sum_{i=1}^N\mathds{1}_{\{Y_i = 1\}}\mathds{1}_{\{L_i \in [\lambda_{j_1}, \lambda_{j_2}]\}}}{\sum_{i=1}^N\mathds{1}_{\{L_i \in [\lambda_{j_1}, \lambda_{j_2}]\}}} - \frac{\sum_{i=1}^N\mathds{1}_{\{Y_i = 1\}}}{N}.
    \end{gathered}
    \label{eq:psi_approx}
\end{equation}

From Equations \eqref{eq:matrix_u_approx} and \eqref{eq:psi_approx}, we deduce how the values of $\Psi_{j_1, j_2}$ can be computed using dynamic programming strategy.

\RestyleAlgo{ruled}
\SetKwComment{Comment}{/* }{*/}
\begin{algorithm}[h]
\SetKwFunction{IterativelyCount}{count\_on\_all\_ranges}
\caption{
\texttt{count\_on\_all\_ranges(sum\_by\_bin)}: Iteratively counting the number of observations $\forall\; [\lambda_{j_1}, \lambda_{j_2}]$, where $j_1 < j_2$, $j_1 = 0,\dots,M{-}1$, $j_2 = 1,\dots,M$
}
    \label{alg:iteratively_count}
    \KwData{$sum\_by\_bin$ - 1D-array of size $M$, representing $\sum_{L \in [\lambda_{j_1}, \lambda_{j_1+1}]} Z$ for observations $L, Z$ (user-defined values)} 
    \KwResult{$sum\_inside\_range$ - 2D-array of size $M \times M$, representing $\sum_{L \in [\lambda_{j_1}, \lambda_{j_2}]} Z$ for observations $L, Z$}
    \BlankLine
    $sum\_inside\_range \gets [[\dots]]$     
    \Comment*[r]{Initialize array}
    $sum\_inside\_range[0][0] \gets sum\_by\_bin[0]$\;
    \For{$j_2 \gets 1 \text{ to } M-1$ }{
        $sum\_inside\_range[0][j_2] \gets sum\_inside\_range[0][j_2-1] + sum\_by\_bin[j_2]$\;
    }
    \For{$j_1 \gets 1 \text{ to } M-1$}{
        \For{$j_2 \gets j_1 \text{ to } M-1$}{
            $sum\_inside\_range[j_1][j_2] \gets sum\_inside\_range[j_1-1][j_2] - sum\_by\_bin[j_1-1]$\;
        }
    }
    \Return {$sum\_inside\_range$}
\end{algorithm}

\RestyleAlgo{ruled}
\SetKwComment{Comment}{/* }{*/}
\SetKwFunction{IterativelyCount}{count\_on\_all\_ranges}

\begin{algorithm}[h]
    \caption{Calculating values of $\Psi_{j_1, j_2}$ for all possible intervals $[\lambda_{j_1}, \lambda_{j_2}]$, given the grid $\lambda_0 < \lambda_1 < ... < \lambda_M$}
    \label{alg:calculate_triangle}
    \KwData{$Y$, $L$ - two 1D-arrays of size $N$ representing observations of random variables \\ \quad \quad \quad Grid of size $M$ given by increasing values $\lambda_0, ..., \lambda_M$ }
    \KwResult{$phi\_by\_range$ - 2D-array of size $M\times M$, representing $U_{\Psi_\Lambda}$}
    \BlankLine
    \BlankLine
    $N \gets size(Y)$ \Comment*[r]{Total number of observations}
    $N\_positive \gets \sum_{i=1}^N \mathds{1}_{\{Y_i=1\}}$ \Comment*[r]{Total number of positive observations}      
    $nb\_obs\_by\_bin \gets [\dots]$ \Comment*[r]{Initialize array}
    $nb\_pos\_obs\_by\_bin \gets [\dots]$ \Comment*[r]{Initialize array}
    \For{$j_1 \gets 1 \text{ to } M$}{
        $nb\_obs\_by\_bin[j_1]$ $\gets$ $\sum_{i=1}^N \mathds{1}_{\{L_i \in [\lambda_{j_1-1}, \lambda_{j_1}]\}}$\; 
        $nb\_pos\_obs\_by\_bin[j_1]$ $\gets$ $\sum_{i=1}^N \mathds{1}_{\{L_i \in [\lambda_{j_1-1}, \lambda_{j_1}]\}}\mathds{1}_{\{Y_i = 1\}}$\;
    }
    $nb\_obs\_inside\_range \gets \IterativelyCount{nb\_obs\_by\_bin}$\;
    $nb\_pos\_obs\_by\_range \gets \IterativelyCount{nb\_pos\_obs\_by\_bin}$\;
    \Comment{Array-wise operations below}
    $proba\_by\_range \gets \frac{nb\_pos\_obs\_inside\_range}{nb\_obs\_inside\_range}$\;
    $phi\_by\_range \gets proba\_by\_range - \frac{N_{positive}}{N}$\;
    \Return {$phi\_by\_range$}
\end{algorithm}

\section{Partition of $L$ on synthetic dataset}\label{appendix:synthetic_data_gaussian}

\begin{figure}[h]
\centering
    \begin{subfigure}[t]{0.49\linewidth}
        \centering
        \includegraphics[width=\linewidth]{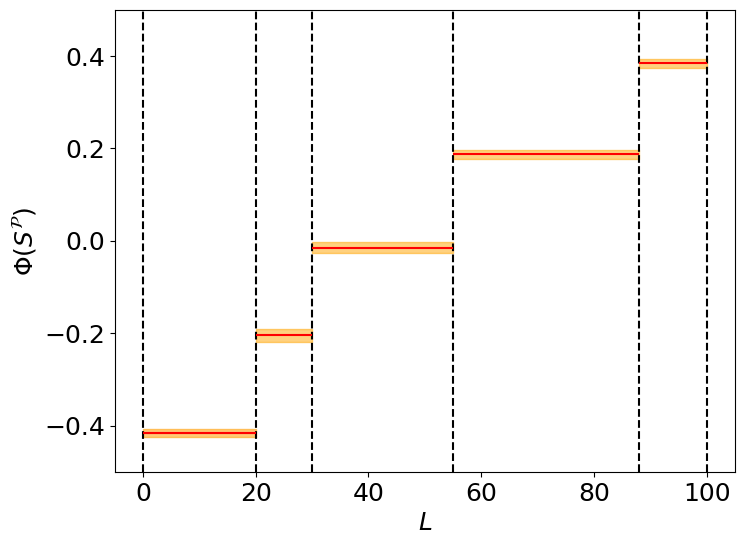}  
        \label{fig:ground_truth_uniform}
    \end{subfigure}
    \hfill
    \begin{subfigure}[t]{0.49\linewidth}
        \centering
        \includegraphics[width=\linewidth]{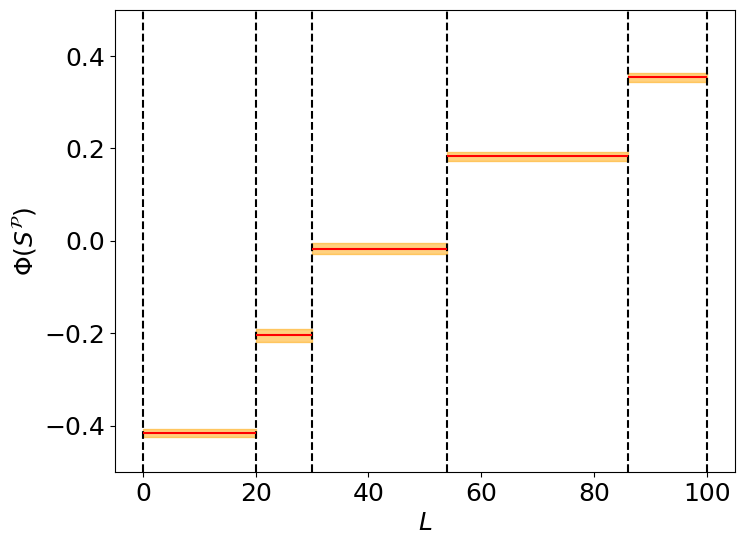}  
    \end{subfigure}
\caption{Partitions on synthetic dataset with uniform distribution of $L$. Ground truth partition (left). K-Means partition (right). \label{fig:partitions_synthetic_data_uniform_p2}}
\vspace{1.5em}
\end{figure}

Figure \ref{fig:partitions_synthetic_data_uniform_p2} shows partitions of $L \sim \mathcal{U}(0,100)$ on synthetic data.

\begin{figure}[h]
\centering
    \begin{subfigure}{\linewidth}
        \centering
        \includegraphics[width=0.5\linewidth]{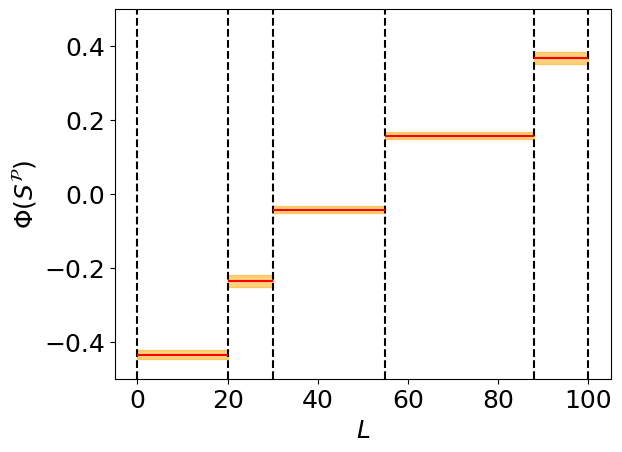}  
    \end{subfigure}
    \begin{subfigure}{0.49\linewidth}
        \centering
        \includegraphics[width=\linewidth]{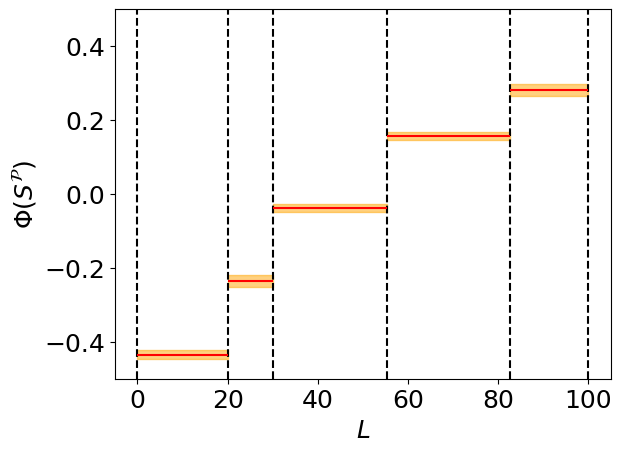}  
    \end{subfigure}
    \hfill    
    \begin{subfigure}{0.49\linewidth}
        \centering
        \includegraphics[width=\linewidth]{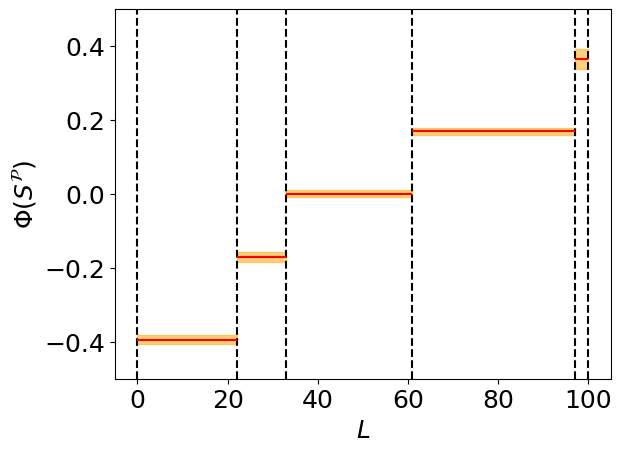}  
    \end{subfigure}
\caption{Partitions on synthetic dataset with truncated Gaussian distribution of $L$. Ground truth partition of $L$ (top). FairGroups partition of $L$ (bottom left). K-Means partition of $L$ (bottom right).
}
\vspace{1.5em}
\label{fig:partitions_synthetic_data_gaussian}
\end{figure}

Figure \ref{fig:partitions_synthetic_data_gaussian} shows partitions of $L \sim \mathcal{N}(50, 20)$ truncated on $[0,100]$ on synthetic data.

\begin{figure}[h]
\centering
    \begin{subfigure}{0.47\linewidth}
        \centering
        \includegraphics[width=\linewidth]{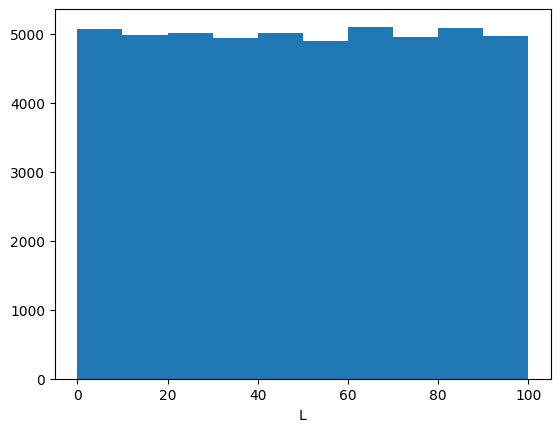}  
        \label{fig:synthetic_L_uniform}
    \end{subfigure}
    \hfill
    \begin{subfigure}{0.49\linewidth}
        \centering
        \includegraphics[width=\linewidth]{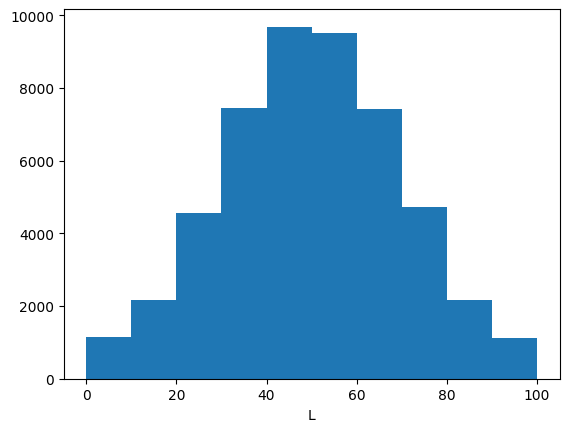}  
        \label{fig:synthetic_L_truncnormal}
    \end{subfigure}
\caption{Synthetic datasets. Generated observations, $L \sim \mathcal{U}(0,100)$ (left). Generated observations, $L \sim \mathcal{N}(50,20)$ truncated on $[0, 100]$ (right).
\label{fig:synthetic_data}
}
\vspace{1.5em}
\end{figure}

\section{Partitions of skin tone in two groups}
We start by comparing partitions of a one-dimensional skin tone variable $L$ with the "predefined" groups following \cite{thong2023beyond}. In fairness analyses, $L$ is usually divided into two groups using the default threshold from \cite{ly2020research}. When $L > 60$, it corresponds to a light skin tone, and conversely, to a dark skin tone when $L \leq 60$.

\begin{figure}[h]
\centering
    \begin{subfigure}{\linewidth}
        \centering
        \includegraphics[width=0.5\linewidth]{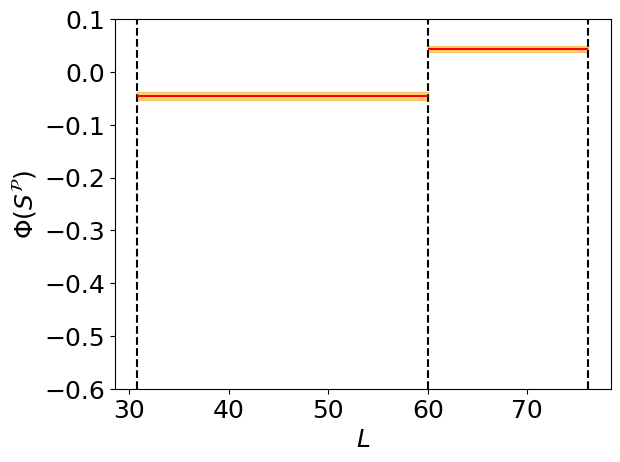}  
        \label{fig:celeba_l_gt}
    \end{subfigure}
    \begin{subfigure}{0.49\linewidth}
        \centering
        \includegraphics[width=\linewidth]{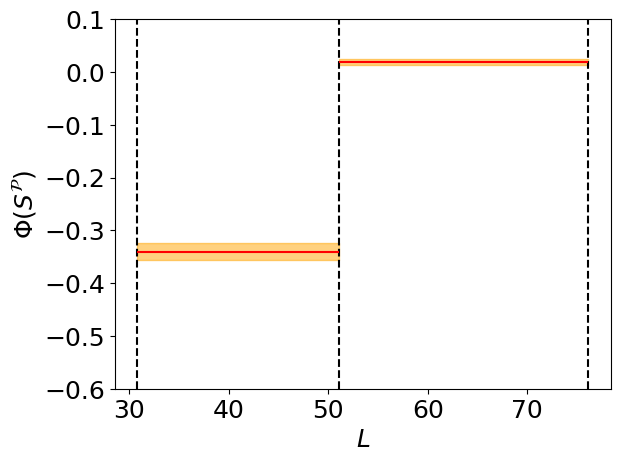}  
        \label{fig:celeba_l_general}
    \end{subfigure}
    \hfill
    \begin{subfigure}{0.49\linewidth}
        \centering
        \includegraphics[width=\linewidth]{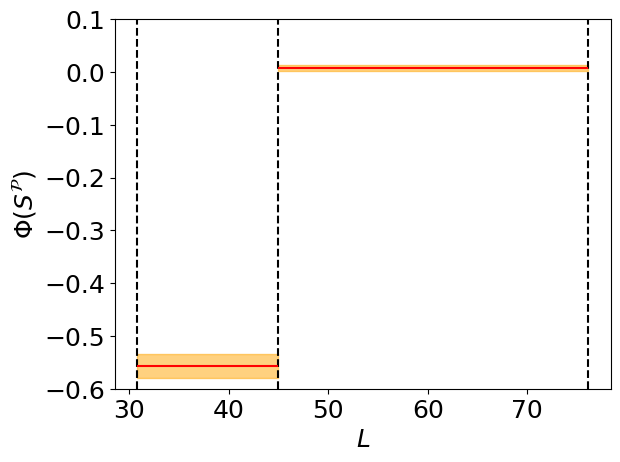}  
        \label{fig:celeba_l_k-means}
    \end{subfigure}
\caption{Partitions of $L$ into the least and the most privileged groups on CelebA. Default partition of $L$ (top). FairGroups partition of $L$ (bottom left). K-Means partition of $L$ (bottom right).
\label{fig:partitions_celeba_L}
}
\vspace{1.5em}
\end{figure}

Figure \ref{fig:partitions_celeba_L} shows the three different partitions of $L$. K-Means and FairGroups partitions identify well the least and the most privileged group. However, K-Means does not account for the impact on the least privileged group. As a result, it produces a smaller least privileged group compared with FairGroups, with a larger confidence interval.

From Figure\ref{fig:partitions_celeba_L}, we conclude that when we default partition $L$ with the threshold used in the literature, we cannot see the discrimination that is present in the model w.r.t. the skin tone. Both groups in the default partition have $\Phi$ close to 0, therefore one might erroneously conclude that the model does not discriminate w.r.t. skin color.

Furthermore, we examine how groups in each partition differ in terms of the level of discrimination. Table \ref{table:L_metrics_k=2} summarizes the metrics for all partitions.

\begin{table*}[htpb]
\centering
\caption{Comparison of partitions of $L$ based on $\Phi(S^\mathcal{P}) = \mathbb{P}(\hat{Y} = 1 | S^\mathcal{P}) - \mathbb{P}(\hat{Y} = 1)$, $K = 2$.}
\label{table:L_metrics_k=2}
\begin{tabular}{|c|c|c|c|c|}
\hline
             Data for computing $\mathcal{P}$
             & \multicolumn{2}{c|}{CelebA} & \multicolumn{1}{c|}{FFHQ} 
             & CelebA \& FFHQ
\\ \hline
\begin{tabular}[c]{@{}c@{}}Metric \\ \end{tabular} & \begin{tabular}[c]{@{}c@{}}$\Var\Big(\Phi\big(S^{\mathcal{P}}\big)\Big)$\\on CelebA\end{tabular} & \begin{tabular}[c]{@{}c@{}}$\Var\Big(\Phi\big(S^{\mathcal{P}}\big)\Big)$\\on FFHQ\end{tabular} & \begin{tabular}[c]{@{}c@{}}$\Var\Big(\Phi\big(S^{\mathcal{P}}\big)\Big)$\\on FFHQ\end{tabular} & \begin{tabular}[c]{@{}c@{}}Rand Index\\ between both partitions,\\ on FFHQ\end{tabular}
\\ \hline
 $\mathcal{P}$: $L \leq 60$ & 0.045 & 0.071 & 0.071 & 1 
\\ \hline
$\mathcal{P}$: K-Means & 0.067 & 0.084 & 0.084 & 0.938                            
\\ \hline
$\mathcal{P}$: FairGroups & \textbf{0.079} & \textbf{0.097} & \textbf{0.101} & 0.945
\\ \hline
\end{tabular}
\end{table*}

Moreover, we measure whether the partitions found for the skin color on CelebA dataset generalize well to the second dataset, FFHQ.
To do so, we first compute the variance $\Var\Big(\Phi(S^\mathcal{P})\Big)$ on FFHQ observations grouped with the partitions $\mathcal{P}$ computed on CelebA. These values can be compared among various partition methods and with the variance for partitions computed on FFHQ, thus more exact.
For each partition method, we finally compare the partitions computed on both dataset using the Rand Index. From the results in Table \ref{table:L_metrics_k=2}, we observe a good generalization of each method since values of $\Var\Big(\Phi(S^\mathcal{P})\Big)$ are close on both datasets. In addition, the Rand Index is closer to 1 for FairGroups, which indicates a higher the stability of the skin tone partition. In these experiments, it means that our methods identify the partitions which reveal how the discrimination is manifested within the space of sensitive attributes, relatively independently of the data distribution.

\section{Debiasing using optimal transport}
\label{appendix:debiasing}

\begin{figure}[h]
    \centering
    \includegraphics[width=0.45\textwidth]{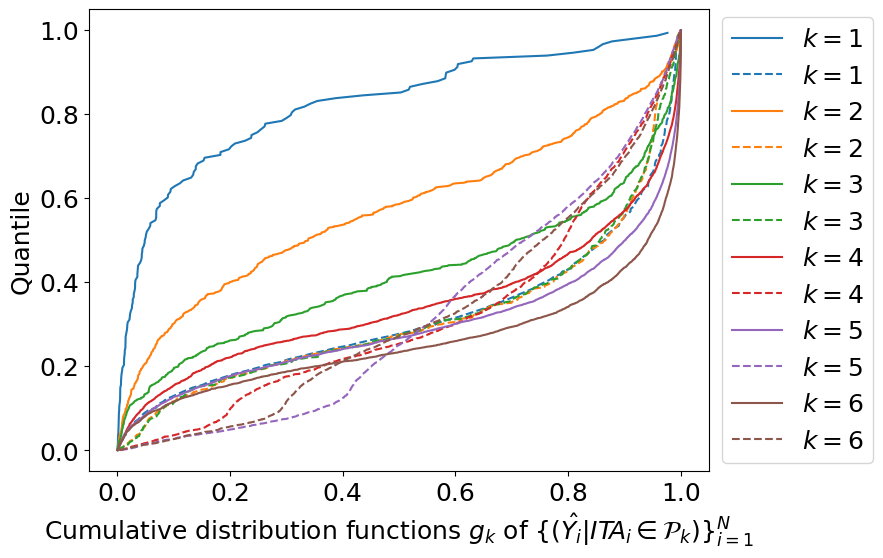}
    \caption{Cumulative distribution functions $g_k$ before (solid lines) and after (dashed lines) post-processing on FairGroups partition of $IT\!A$.}
    \label{fig:debiasing_cdf}
\end{figure}

To demonstrate the mechanism of optimal transport-based post-processing, we operate within the setting described in Section~\ref{section:partitions_ita}, which details the dataset and partitioning scheme used. In this context, we mitigate the bias of the model $g_{\theta}$ using the FairGroups partition. Figure~\ref{fig:debiasing_cdf} illustrates how the cumulative distribution functions (CDFs) of each group $\mathcal{P}_k$ in the partition $\mathcal{P}$ are affected by the post-processing step. Each solid line represents the CDF of a group $\mathcal{P}_k$ prior to post-processing, while the corresponding dashed line depicts the CDF after the application of the post-processing procedure. From Figure~\ref{fig:debiasing_cdf}, we observe that the CDFs across groups become more aligned following post-processing, indicating a reduction in group-level disparities.

However, it is important to note that post-processing does not map all group distributions to a single, identical CDF. Rather, the residual discrepancies between the post-processed CDFs are modulated by the parameter~$\alpha$, which governs the strength of the debiasing and enables a controlled trade-off between predictive accuracy and group fairness.

\vspace{1.0em}

\end{document}